\documentclass[12pt]{article}

\usepackage{scicite}
\usepackage{times}
\usepackage{bbm}
\usepackage{xcolor}
\usepackage{graphicx}
\usepackage{booktabs}
\usepackage{mathtools}
\usepackage{amssymb}
\usepackage{hyperref}
\usepackage{subcaption}
\usepackage{float}
\definecolor{darkblue}{rgb}{0,0.08,0.45}
\hypersetup{%
  pdfborder=0 0 0,
  colorlinks=true,
  citecolor=darkblue,
  urlcolor=darkblue
}

\newcommand{\app}{\raise.17ex\hbox{$\scriptstyle\sim$}}

\newenvironment{sciabstract}{%
\begin{quote} \bf}
{\end{quote}}

\title{Real-World Humanoid Locomotion \\ with Reinforcement Learning}
\author
{Ilija Radosavovic$^\ast$, Tete Xiao$^\ast$, Bike Zhang$^\ast$,\\
Trevor Darrell$^\dagger$, Jitendra Malik$^\dagger$, Koushil Sreenath$^\dagger$\\
\\
\normalsize{University of California, Berkeley}\\
\normalsize{\href{https://learning-humanoid-locomotion.github.io/}{Project Page}}
}
\date{}

\begin{document} 
\baselineskip24pt
\maketitle 

\begin{sciabstract}
Humanoid robots that can autonomously operate in diverse environments have the potential to help address labour shortages in factories, assist elderly at homes, and colonize new planets. While classical controllers for humanoid robots have shown impressive results in a number of settings, they are challenging to generalize and adapt to new environments. Here, we present a fully learning-based approach for real-world humanoid locomotion. Our controller is a causal transformer that takes the history of proprioceptive observations and actions as input and predicts the next action. We hypothesize that the observation-action history contains useful information about the world that a powerful transformer model can use to adapt its behavior in-context, without updating its weights. We train our model with large-scale model-free reinforcement learning on an ensemble of randomized environments in simulation and deploy it to the real world zero-shot. Our controller can walk over various outdoor terrains, is robust to external disturbances, and can adapt in context.
\end{sciabstract}

\newpage
\section*{Introduction}

The dream of robotics has always been that of general purpose machines that can perform many tasks in diverse, unstructured environments. Examples include moving boxes, changing tires, ironing shirts, and baking cakes. This grand goal calls for a general purpose embodiment and a general purpose controller. A humanoid robot could, in principle, deliver on this goal.

Indeed, roboticists designed the first full-sized real-world humanoid robot \cite{Kato1973} in the 1970s. Since then, researchers have developed a variety of humanoid robots to push the limits of robot locomotion research~\cite{Hirai1998,Nelson2012,Stasse2017,Chignoli2021}. The control problem, however, remains a considerable challenge. While classical control methods can achieve stable and robust locomotion \cite{Raibert1986,Kajita2001,Westervelt2003,Collins2005}, optimization-based strategies have shown the advantage of simultaneously authoring dynamic behaviors and obeying constraints \cite{Tassa2012,Kuindersma2016,Di2018}. The most well-known are the examples of the Boston Dynamics Atlas robot doing back flips, jumping over obstacles, and dancing.

While these approaches have made great progress, learning-based methods have become of increasing interest due to their ability to learn from diverse simulations or real environments. For example, learning-based approaches have proven very effective in dexterous manipulation~\cite{Openai2018,Openai2019,handa2023dextreme}, quadrupedal locomotion~\cite{Hwangbo2019,Lee2020,Kumar2021}, and bipedal locomotion~\cite{Benbrahim1997,tedrake2004stochastic,Xie2018,Siekmann2021sim,Siekmann2021blind}. Moreover, learning-based approaches have been explored for small-sized humanoids~\cite{iida2004humanoid,rodriguez2021deepwalk} and combined with model-based controllers for full-sized humanoids~\cite{Castillo2022,Krishna2022} as well.

\begin{figure*}
\centering
\includegraphics[width=\linewidth]{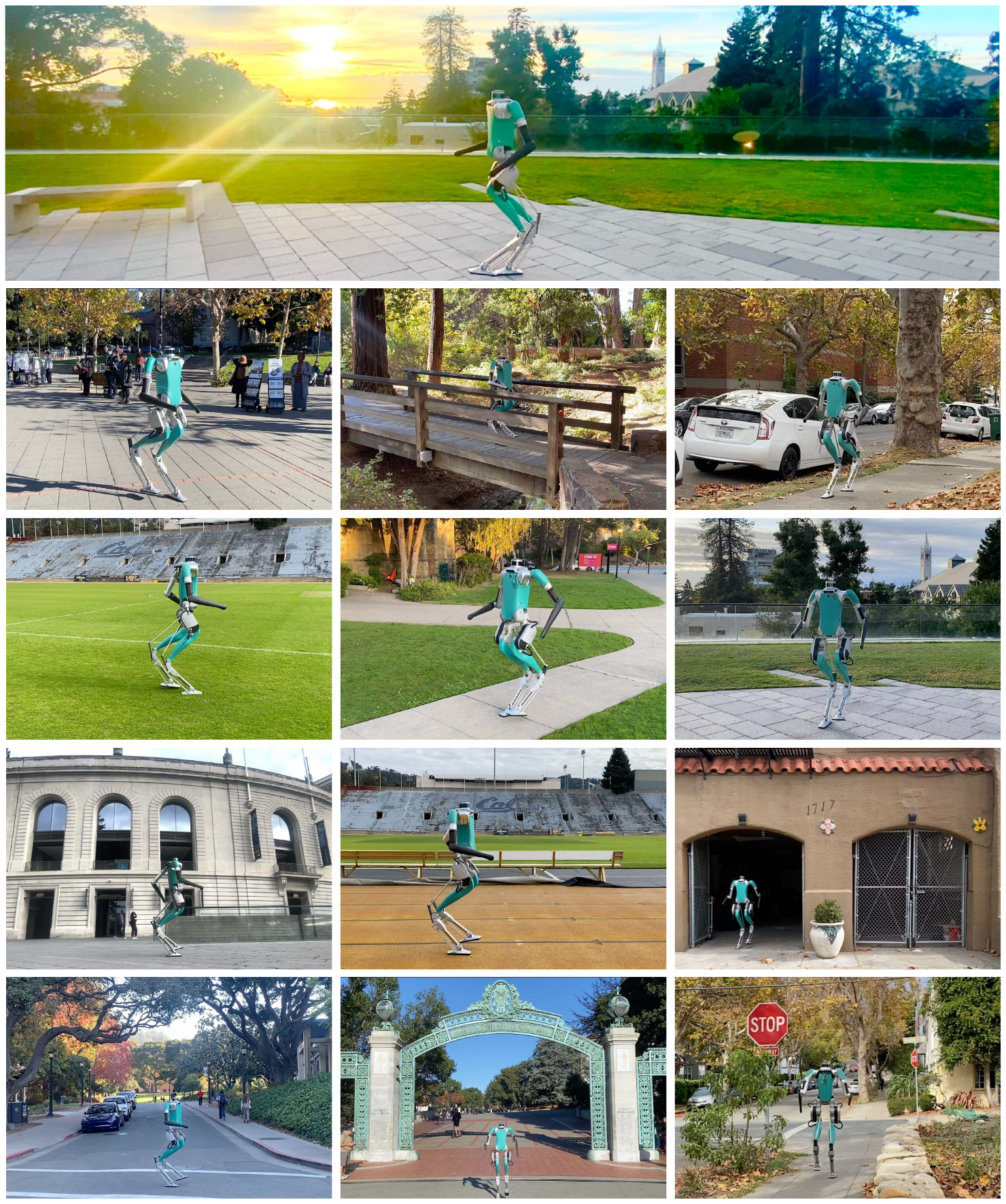}
\caption{\textbf{Deployment to outdoor environments.} We deploy our model to a number of outdoor environments. Example videos are shown in~\href{https://youtu.be/Wd1q8KaNuME}{Movie 1}. We find that our controller is able to traverse a range of everyday environments including plazas, side walks, tracks, and grass fields.}
\label{fig:cover_outdoor}
\end{figure*}

In this paper, we propose a learning-based approach for real-world humanoid locomotion (Figure~\ref{fig:cover_outdoor}). Our controller is a causal transformer that takes the history of proprioceptive observations and actions as input and predicts the next action (Figure~\ref{fig:framework}, C). Our model is trained with large-scale reinforcement learning on thousands of randomized environments in simulation and deployed to the real world in a zero-shot fashion (Figure~\ref{fig:framework}, A-B).

\newpage

Our approach falls in the general family of techniques for sim-to-real transfer with domain randomization~\cite{Antonova2017, Sadeghi2017, Tobin2017, Peng2018}. Among these, the recent approaches for learning legged locomotion have employed either memory-based networks like Long Short-Term Memory (LSTM)~\cite{Openai2019,Siekmann2021blind} or trained an explicit estimator to regress environment properties from Temporal Convolutional Network (TCN) features~\cite{Lee2020,Kumar2021}.

We hypothesize that the history of observations and actions implicitly encodes the information about the world that a powerful transformer model can use to adapt its behavior dynamically at test time. For example, the model can use the history of desired vs actual states to figure out how to adjust its actions to better achieve future states. This can be seen as a form of in-context learning often found in large transformer models like GPT-3~\cite{Brown2020}.

\begin{figure}
\centering
\includegraphics[width=1.0\linewidth]{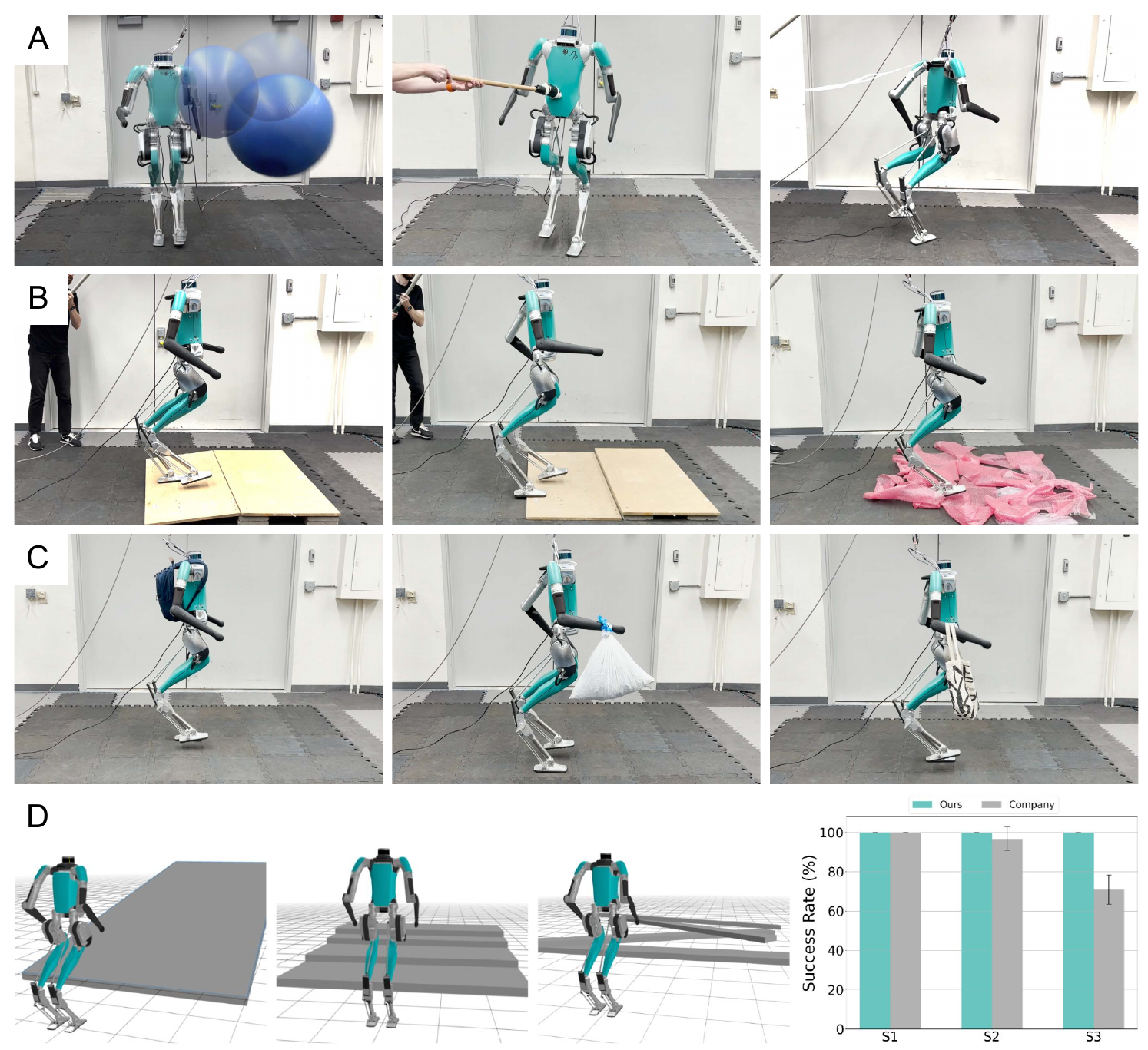}
\caption{\textbf{Indoor and simulation experiments.} We test the robustness of our controller to (\textbf{A}) external disturbances, (\textbf{B}) different terrains, and (\textbf{C}) payloads. Videos are shown in \href{https://youtu.be/cdbWFNvT72c}{Movie 7}. We find that our controller is able to tackle of the scenarios successfully, including those that are considerably out of the training distribution. (\textbf{D}) We find that our controller outperforms the state-of-the-art company controller across three different settings in simulation. The gains are larger for harder terrains, like steps and unstable ground. We replicate a subset of the scenarios on hardware and observe consistent behaviors, which can be seen in examples from~\href{https://youtu.be/MUgey-1j5tE}{Movie 2}.}
\label{fig:indoors}
\end{figure}

We evaluate our model on a full-sized humanoid robot through a series of real-world and simulated experiments. We show that our policy enables reliable outdoor walking without falls (Figure~\ref{fig:cover_outdoor}), is robust to external disturbances, can traverse different terrains, and carry payloads of varying mass (Figure~\ref{fig:indoors}A-C). Moreover, we find that our approach compares favorably to the state-of-the-art model-based controller (Figure~\ref{fig:indoors}D). Our policy exhibits natural walking behaviors, including following different commands (Figure~\ref{fig:natural_walking}), high-speed locomotion, and an emergent arm swing motion (Figure~\ref{fig:natural_walking_plot}). Importantly, our policy is adaptive and can change its behavior based on context, including gradual gait changes based on slowly varying terrains (Figure~\ref{fig:adaptation_1}) and rapid adaptation to sudden obstacles (Figure~\ref{fig:adaptation_2}). To understand different design choices, we analyze our method in controlled experiments and find that the transformer architecture outperforms other neural network architectures, the model benefits from larger context, and that joint training with teacher imitation and reinforcement learning is beneficial (Figure~\ref{fig:ablations}).

Our results suggest that simple and general learning-based controllers are capable of complex, high-dimensional humanoid control in the physical world. We hope that our work will encourage future exploration of scalable learning-based approaches for humanoid robotics.

\section*{Results}

\paragraph*{Digit humanoid robot.} Digit is a general-purpose humanoid robot developed by Agility Robotics, standing at approximately 1.6 meters tall with a total weight of 45 kilograms. The robot's floating-base model is equipped with 30 degrees of freedom, including four actuated joints in each arm and eight joints in each leg, of which six are actuated. The passive joints, the shin and tarsus, are designed to be connected through the use of leaf springs and a four-bar linkage mechanism, while the toe joint is actuated by means of rods attached at the tarsus joint. Digit robot has been used as a humanoid platform for mechanical design~\cite{Han2022}, locomotion control~\cite{Castillo2021,Krishna2022,Gao2022time}, state estimation~\cite{Gao2022invariant}, planning~\cite{Adu2022,Narkhede2022,shamsah2023integrated}, etc.

\subsection*{Outdoor deployment}

We begin by reporting the results of deploying our controller to a number of outdoor environments. Examples are shown in Figure~\ref{fig:cover_outdoor} and~\href{https://youtu.be/Wd1q8KaNuME}{Movie 1}. These include everyday human environments, plazas,  walkways, sidewalks, running tracks, and grass fields. The terrains vary considerably in terms of material properties, like concrete, rubber, and grass, as well as conditions, like dry in a sunny afternoon or wet in the early morning. Our controller is trained entirely in simulation and deployed to the real world zero-shot. The terrain properties found in the outdoor environments were not encountered during training. We found that our controller was able to walk over all of the tested terrains reliably and were comfortable deploying it without a safety gantry. Indeed, over the course of one week of full-day testing in outdoor environments we did not observe any falls. Nevertheless, since our controller acts based on the history of observations and actions and does not include any additional sensors like cameras, it can bump and get trapped by obstacles like steps, but manage to adapt its behavior to avoid falling (see Section~\ref{fig:adaptation_2} for additional discussion and analysis of adaptation).

\subsection*{Indoor and simulation experiments}

We conduct a series of experiments in the laboratory environment to test the performance of the proposed approach in controlled settings (Figure~\ref{fig:indoors}).

\paragraph*{External forces.} Robustness to external forces is a critical requirement for real-world deployment of humanoid robots. We test if our controller is able to handle sudden external forces while walking. These experiments include throwing a large yoga ball at the robot, pushing the robot with a wooden stick, and pulling the robot from the back while it is walking forward (Figure~\ref{fig:indoors}A). We find that our controller is able to stabilize the robot in each of these scenarios. Given that the humanoid is a highly unstable system and that the disturbances we apply are sudden, the robot must react in fractions of a second and adjust its actions to avoid falling.

\paragraph*{Rough terrain.} In addition to handling external disturbances, a humanoid robot must also be able to locomote over different terrains. To assess the capabilities of our controller in this regard, we conduct a series of experiments on different terrains in the laboratory (Figure~\ref{fig:indoors}B). Each experiment involved commanding the robot to walk forward at a constant velocity of 0.15 m/s. Next, we covered the floor with four different types of items: rubbers, cloths, cables, and bubble wraps, which altered the roughness of the terrain and could potentially lead to challenging entanglement and slipping situations, as the robot does not utilize exteroceptive sensing. Despite these impediments, our controller was able to traverse all these terrain types. Finally, we evaluated the controller's performance on two different slopes. Our simulations during training time included slopes up to 10\% grade, and our testing slopes are up to 8.7\% grade. Our results demonstrate that the robot was able to successfully traverse both slopes, with more robustness at higher velocity (0.2 m/s) on steeper slopes.

\paragraph*{Payloads.} Next, we evaluate the robot's ability to carry loads of varying mass, shape, and center-of-mass while walking forward (Figure~\ref{fig:indoors}C). We conduct five experiments, each with the robot carrying a different type of load: an empty backpack, a loaded backpack, a cloth handbag, a loaded trash bag, and a paper bag. Our results demonstrate that the robot is able to successfully complete its walking route while carrying each of these loads. Notably, our learning-based controller is able to adapt to the presence of a loaded trash bag attached to its arm, despite the reliance of our policy on arm swing movements for balancing. This suggests that our controller is able to adapt its behavior according to the context.

\paragraph*{Comparison to the state of the art.} We compare our controller to the company controller provided by Agility Robotics, which is the state of the art for this robot. To quantify the performance across many runs, we use the high-fidelity simulator by Agility Robotics. We consider three different scenarios: walking over slopes, steps, and unstable ground (Figure~\ref{fig:indoors}D). We command the robot to walk forward and consider a trial as successful if the robot crosses the terrain without falling. Crossing a portion of the terrain obtains partial success. We report the mean success rate with 95\% CI per terrain across 10 runs (Figure~\ref{fig:indoors}D). We find that both ours and the company controller work well on slopes. Next, we see that our controller outperforms the company controller on steps. The company controller struggles to correct itself from foot trapping and shuts off. We replicated this scenario in the real world and have observed consistent behavior, shown in~\href{https://youtu.be/MUgey-1j5tE}{Movie 2}. In contrast, our controller is able to recover successfully. Note that our controller was \emph{not} trained on steps in simulation and that the foot-trapping recovery behaviors are emergent (see also Section~\ref{fig:adaptation_2}). Finally, we compare the two controllers on a terrain with unstable planks. This setting is challenging as the terrain can dislodge under the robot feet. We find that our controller considerably outperforms the company controller. We did not evaluate the controllers on this terrain in the real world due to concerns for potential hardware damage.

\subsection*{Natural walking}

\begin{figure}
\centering
\includegraphics[width=1.0\linewidth]{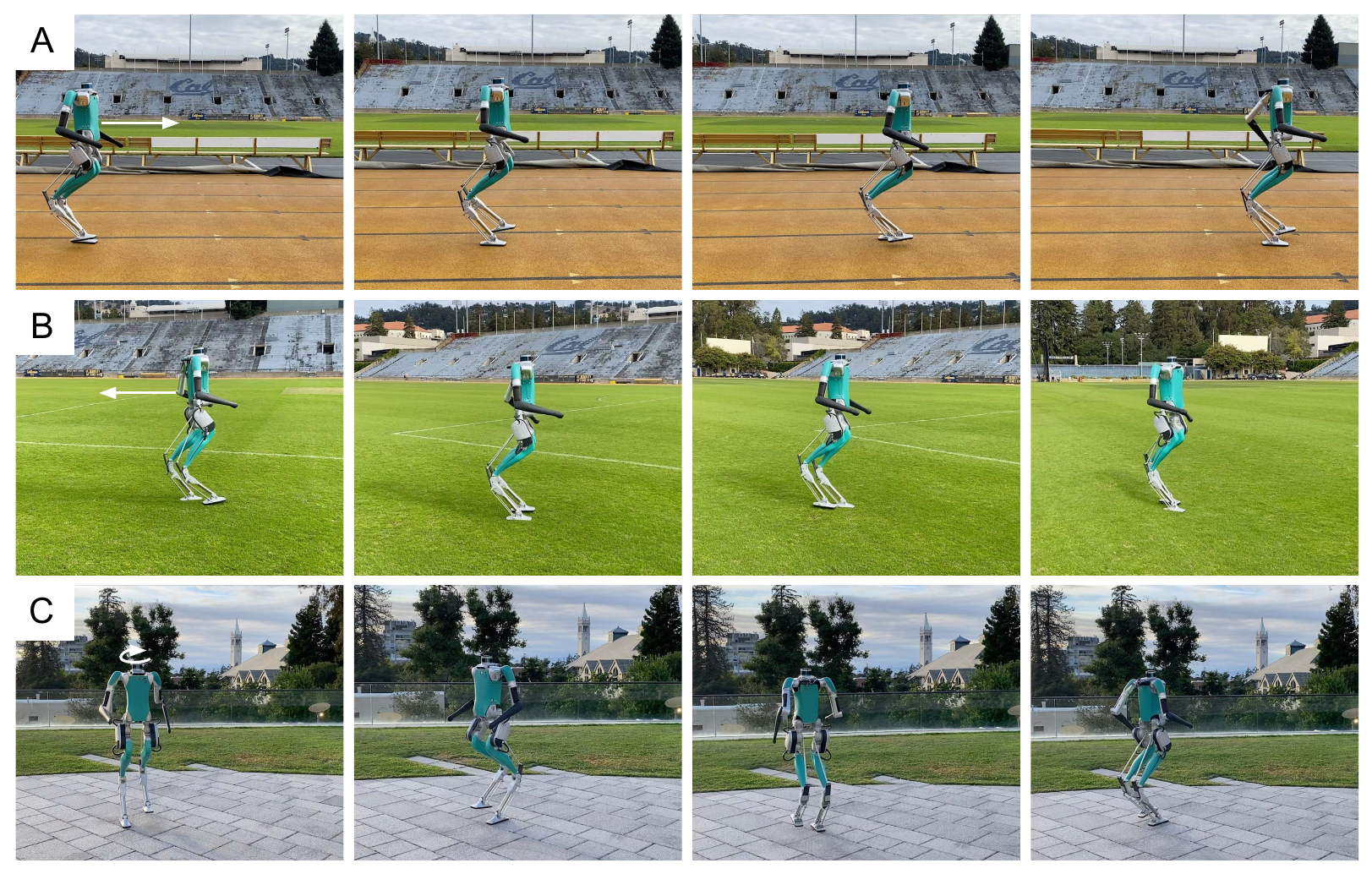}
\caption{\textbf{Omnidirectional walking.} Our learning-based controller is able to accurately follow a range of velocity commands to perform omni-directional locomotion, including (\textbf{A}) walking forward, (\textbf{B}) backward, and (\textbf{C}) turning. Video examples are shown in~\href{https://youtu.be/7bChPZWTAig}{Movie 3}.}
\label{fig:natural_walking}
\end{figure}

\paragraph*{Omnidirectional walking.} Our controller performs omnidirectional locomotion by following velocity commands. Specifically, it is conditioned on linear velocity on the x-axis, linear velocity on the y-axis, and angular velocity around the z-axis. At training time, we sample commands randomly every 10 seconds (see the Appendix for details). At deployment, we find that our controller is able to follow commands accurately. In addition, it generalizes to continuously changing commands, supplied via a joystick in real time, which is different from training. We show examples of walking forward, backward, and turning in Figure~\ref{fig:natural_walking}, and in~\href{https://youtu.be/7bChPZWTAig}{Movie 3}.

\begin{figure}
\centering
\includegraphics[width=1.0\linewidth]{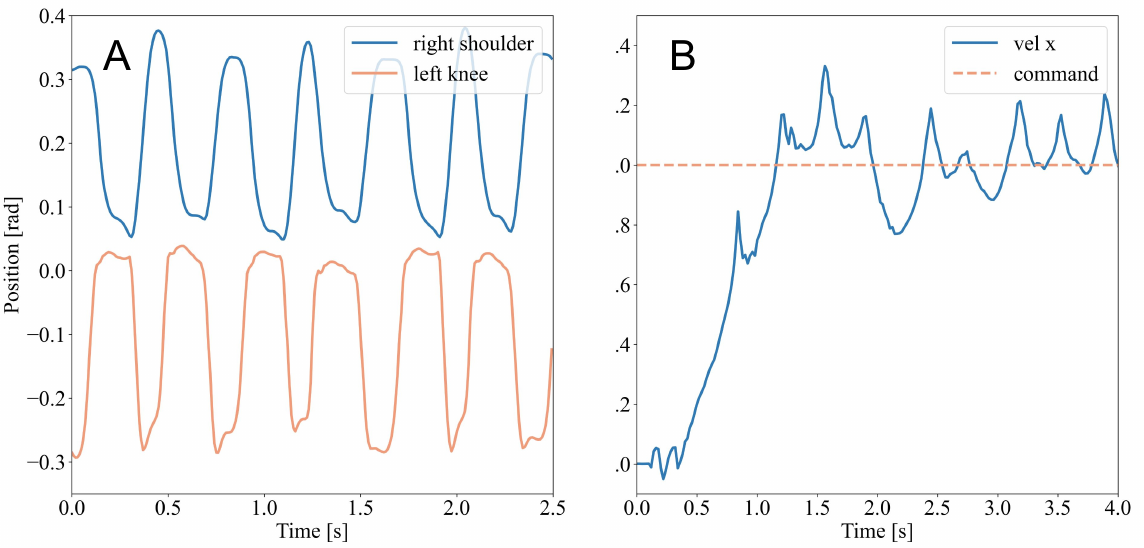}
\caption{\textbf{Arm swing and fast walking.} (\textbf{A}) The learned humanoid locomotion in our experiments exhibits human-like arm swing behaviors in coordination with leg movements, i.e., a contralateral relationship between the arms and the legs. (\textbf{B}) Our controller is able to perform fast walking on hardware. The video is shown in ~\href{https://youtu.be/gD9Y-hvfBic}{Movie 4}.}
\label{fig:natural_walking_plot}
\end{figure}

\paragraph*{Dynamic arm swing.} A distinct feature of natural human walking is the arm swing. Studying the arm-swing behavior in humans has a very long history in biomechanics~\cite{morton1952human, herr2008angular, Collins2009}. There are a number of existing hypothesis for why humans might be swinging their arms while walking. Examples include that arm-swinging leads to dynamic stability~\cite{ortega2008effects}, that it reduces a metabolic energy cost of walking~\cite{umberger2008effects}, and that it is an ancestral trait conserved from quadrupedal coordination~\cite{murray1967patterns}. We are particularly inspired by the work of~\cite{Collins2009}, which suggests that arm swinging may require little effort while providing substantial energy benefit. We test this hypothesis empirically during multiple types of arm motion including swinging, arms bound or held to the body, and arms swinging with phase opposite to normal.

When training our neural network controller, we do not impose explicit constraints on the arm swing motion in the reward function or use any reference trajectories to guide the arm motions. Interestingly, we observe that our trained policy exhibits an emergent arm swing behavior similar to natural human walking, as shown in Figure~\ref{fig:natural_walking_plot}A. The swinging arm is coordinated with the legs like humans. Specifically, when the left leg is lifting up, the right arm swings forward. We note that our reward function includes energy minimization terms which might suggest that the emergent arm swing motion might lead to energy savings in humanoid locomotion as well.

\begin{figure}
\centering
\includegraphics[width=1.0\linewidth]{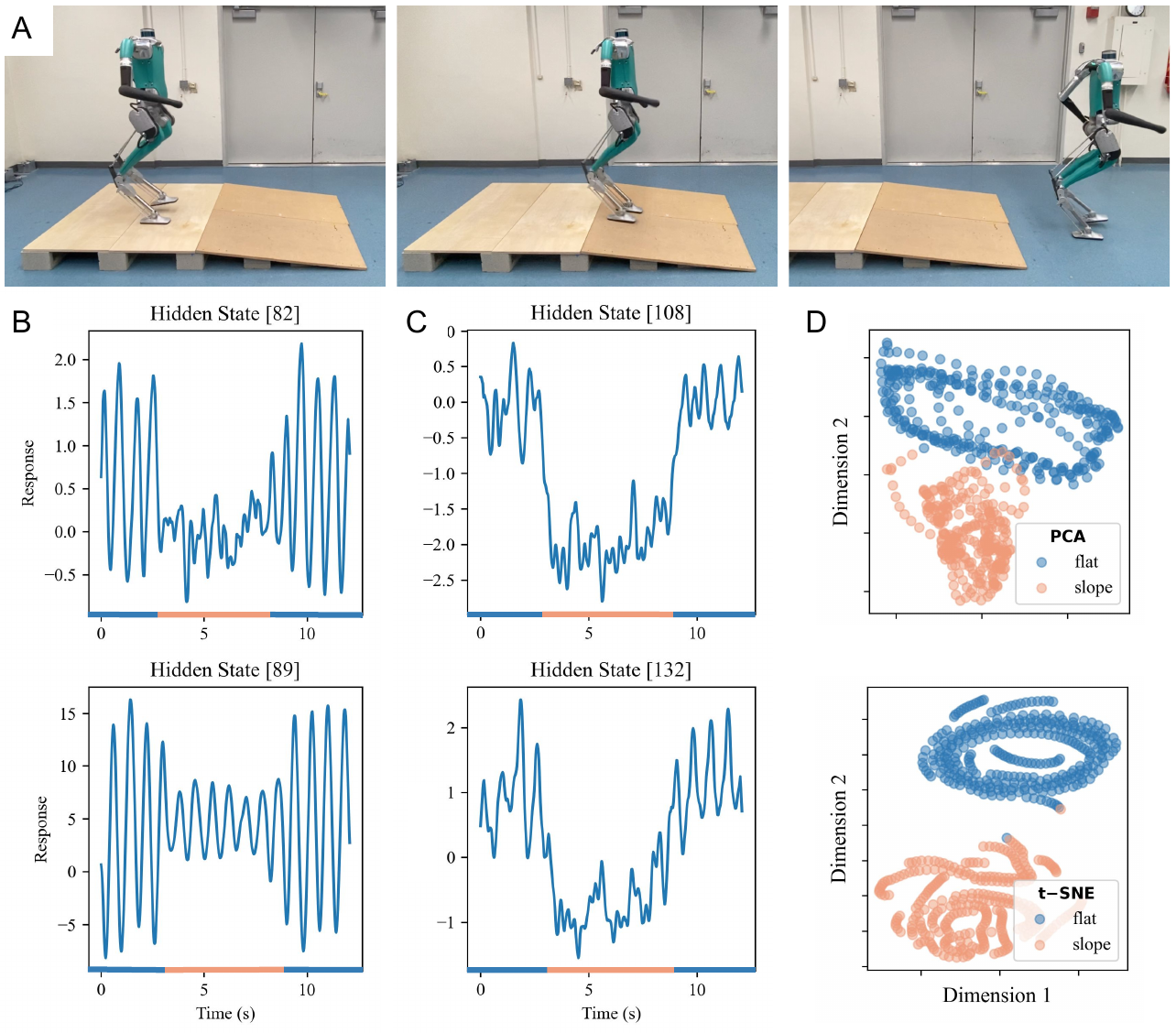}
\caption{\textbf{Gait changes based on terrain type.} (\textbf{A}) We command the robot to walk forward over a course consisting of three sections: flat, downward slope, and flat again. We observe that our controller adapts its behavior based on terrain, changing the gait from natural walking on flat terrain, to small steps on downward slope, to natural walking on flat terrain again. Video is shown in~\href{https://youtu.be/ByEk-D3TevM}{Movie 5}. This type of adaptation based on context is emergent and has not been pre-specified during training. (\textbf{B}) We analyze the hidden state of the last layer of our neural network controller and find that certain neuron responses correlate with the gait patterns observed over different terrain sections. (\textbf{C}) In addition, some of the neuron responses correlate changes in the terrain and are high for flat sections and low for the slope section. (\textbf{D}) To analyze the neural responses in aggregate, we project the 192-dimensional hidden states to two dimensions using PCA and t-SNE. Each data point corresponds to one timestep and is color-coded by the terrain section. We see that the hidden states get grouped into clear clusters based on the terrain type.}
\label{fig:adaptation_1}
\end{figure}

\paragraph*{Fast walking.} There is a considerable difference between walking at low and high speeds. We analyze the performance of our controller when walking fast in the real world. In Figure~\ref{fig:natural_walking_plot}B, we show the velocity tracking performance given a commanded step velocity at 1 m/s. The corresponding video is in \href{https://youtu.be/gD9Y-hvfBic}{Movie 4}. We see that the robot is able to achieve the commanded velocity from rest within 1 s and track it accurately for the duration of the course.

\subsection*{In-context adaptation}\label{sec:adaptation}

\paragraph*{Emergent gait changes based on terrain.} We command the robot to walk forward over a terrain consisting of three sections in order: flat ground, downward slope, and flat ground, shown in Figure~\ref{fig:adaptation_1}A. We find that our controller changes its walking behavior entirely based on the terrain. Specifically, it starts by normal walking on flat ground, transitions to using small steps without lifting its legs much on downward slope, and back to normal walking on flat ground again. These behavior changes are emergent and were not pre-specified.

To understand this behavior better, we study the patterns of neural activity of our transformer model over time. First, we look at the responses of individual neurons. We find that certain neurons correlate with gait. Namely, they have high amplitude during walking on flat and low amplitude on the downward slope. Two such neurons are shown in Figure~\ref{fig:adaptation_1}B. Moreover, some neurons correlate with terrain types. Their responses are high on flat terrain and low on slope, as shown in Figure~\ref{fig:adaptation_1}C. We also analyze the neural responses in aggregate by performing dimensionality reduction. We project the 192-dimensional hidden state from each timestep into a 2-dimensional vector using PCA and t-SNE. In Figure~\ref{fig:adaptation_1}D, we show the results color-coded by terrain type (terrain labels only used for visualization) and see clear clusters based on terrain. These suggest that our representations capture important terrain and gait related properties.

\begin{figure}[t]
\centering
\includegraphics[width=1.0\linewidth]{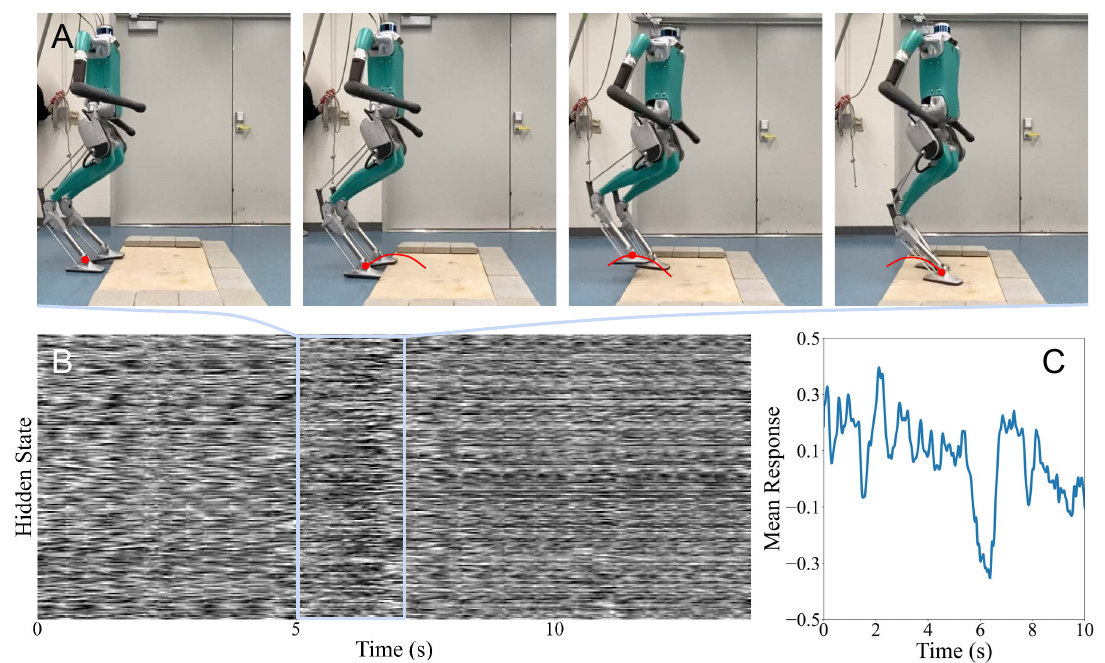}
\caption{\textbf{Emergent recovery from foot-trapping.} (\textbf{A}) Our controller is able to adapt to discrete obstacles not seen during training and recovers from foot-trapping by lifting its legs higher and faster on subsequent attempts. This behavior is consistent and representative examples are shown in~\href{https://youtu.be/QLdX6o2RMcU}{Movie 6}. (\textbf{B}) We analyze the hidden state of the last layer of our transformer model and find that there is a change in the pattern of activity that correlates with the foot-trapping events. (\textbf{C}) Mean activation responses contain clear spikes during foot-trapping events as well.}
\label{fig:adaptation_2}
\end{figure}

\paragraph*{Emergent recovery from foot-trapping.} Next, we study the ability of our controller to recover from foot-trapping that occurs when one of the robot legs hits a discrete step obstacle. Note that steps or other form of discrete obstacles were not seen during training. This setting is relevant since our robot is blind and may find itself in such situations during deployment. We find that our controller is still able to detect and react to foot-trapping events based on the history of observations and actions. Specifically, after hitting the step with its leg the robot will attempt to lift its legs higher and faster on subsequent attempts. Figure~\ref{fig:adaptation_2}A, shows an example episode. We show a representative example for one of each of the two legs in~\href{https://youtu.be/QLdX6o2RMcU}{Movie 6}. We find that our controller is able to recover from different variations of such scenarios consistently. This behavior is emergent and was not pre-programmed or encouraged during training.

To understand this behavior better, we study the pattern of neural activity during an episode that contains foot-trapping and recovery, shown in Figure~\ref{fig:adaptation_2}. In Figure~\ref{fig:adaptation_2}B, we plot the neural activity over time. Each column is a 192-dimensional hidden state of the last layer of our transformer model and each row is the value of an individual neuron over time. We see a clear change in the pattern in activity, highlighted with a rectangle, that occurs during the foot-trapping event. In Figure~\ref{fig:adaptation_2}C, we show the mean neuron response over time and see that there is a clear deviation from normal activity during the foot-trapping event. These suggest that our transformer model is able to implicitly detect such events based on neural activity.

\section*{Discussion}

We present a learning-based controller for full-sized humanoid locomotion. Our controller is a causal transformer that takes the history of past observations and actions as input and predicts future actions. We train our model using large-scale simulation and deploy it to the real world in a zero-shot fashion. We show that our policy enables reliable outdoor walking without falls, is robust to external disturbances, can traverse different terrains, and carry payloads of varying mass. Our policy exhibits natural walking behaviors, including following different commands, high-speed locomotion, and an emergent arm swing motion. Moreover, we find that our controller can adapt to novel scenarios at test time by changing its behavior based on context, including gait changes based on the terrain and recovery from foot-trapping.

\paragraph*{Limitations.} Our approach shows promising results in terms of adaptability and robustness to different terrains and external disturbances. However, it still has some limitations that need to be addressed in future work. One limitation is that our policy is not perfectly symmetrical, as the motors on two sides do not produce identical trajectories. This results in a slight asymmetry in movement, with the controller being better at lateral movements to the left compared to the right. Additionally, our policy is not perfect at tracking the commanded velocity. Finally, under excessive external disturbances, like a very strong pull of a cable attached to the robot, can cause the robot to fall.

\paragraph*{Possible extensions.} Our neural network controller is a general transformer model. Compared to alternate model choices, like TCN and LSTM, this has favorable properties that can be explored in future work. For example, it should be easier to scale with additional data and compute~\cite{Kaplan2020} and enable us to incorporate additional input modalities~\cite{Alayrac2022}. Analogous to fields like vision~\cite{Dosovitskiy2021} and language~\cite{Radford2018}, we believe that transformers may facilitate our future progress in scaling learning approaches for real-world humanoid locomotion.

\section*{Materials and Methods}\label{sec:method}

This section describes in detail the policy learning procedure, the simulation process, the sim-to-real transfer deployment, and the analysis of the transformer-based controller. An overview of our method is shown in Figure~\ref{fig:framework}. The policy learning includes two steps: teacher state policy training and student observation policy learning. We adopt a massively parallel simulation environment, where we introduce a simulation method that can simulate closed kinematic chains enabling us to simulate the underactuated Digit humanoid robot. We explain the procedure for sim-to-real transfer in detail. Finally, we provide analysis of our transformer policy.

\subsection*{Policy learning}

\paragraph*{Problem formulation.}

We formulate the control problem as a Markov Decision Process (MDP), which provides a mathematical framework for modeling discrete-time decision-making processes. The MDP comprises the following elements: a state space $S$, an action space $A$, a transition function $P(s_{t+1}|s_{t}, a_{t})$ that determines the probability of transitioning from state $s_{t}$ to $s_{t+1}$ after taking action $a_{t}$ at time step $t$, and a scalar reward function $R(s_{t+1}|s_{t}, a_{t})$, which assigns a scalar value to each state-action-state transition, serving as feedback to the agent on the quality of its actions. Our approach to solving the MDP problem is through Reinforcement Learning (RL), which aims to find an optimal policy that maximizes the expected cumulative reward over a finite or infinite horizon.

\begin{figure}
\centering
\includegraphics[width=1.0\linewidth]{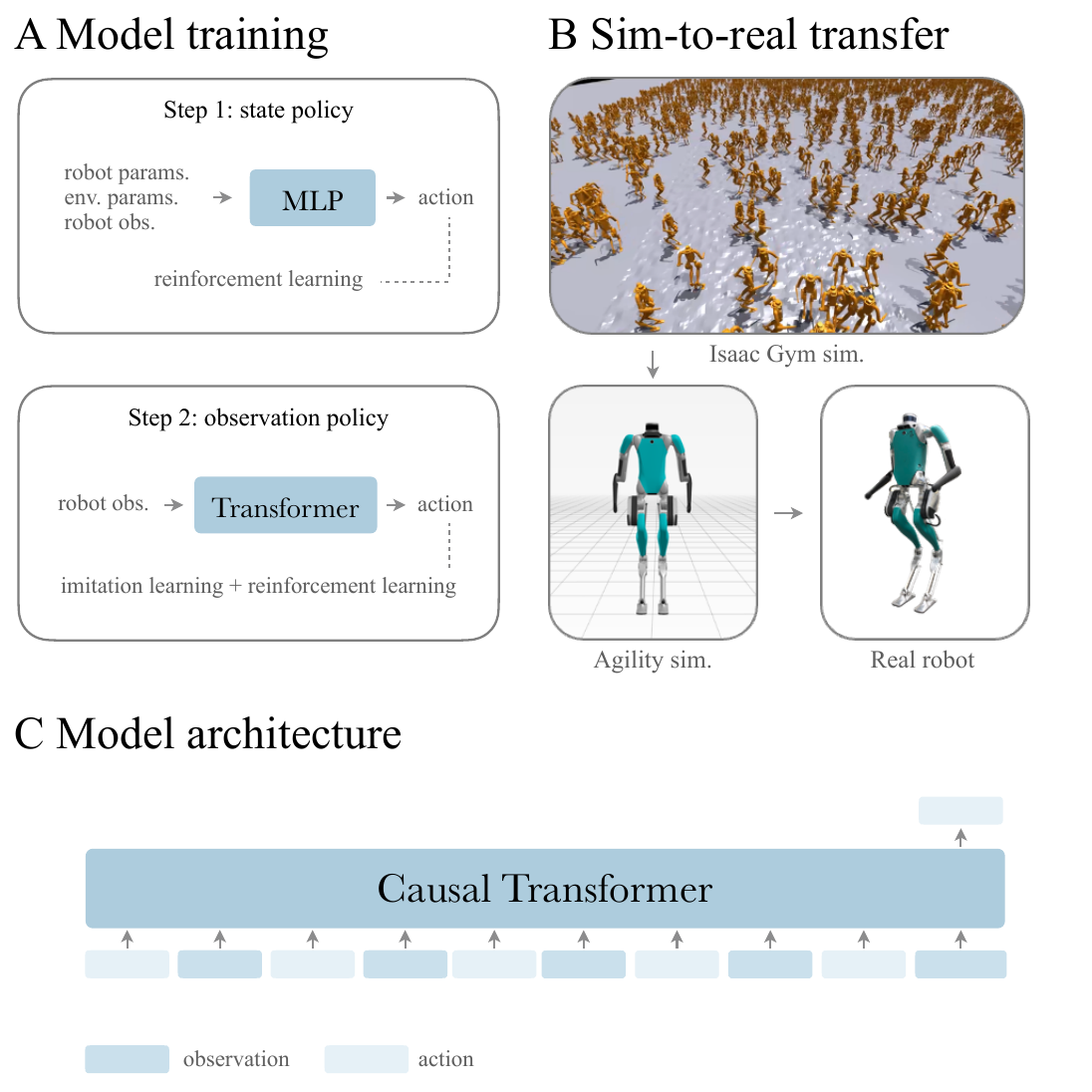}
\caption{\textbf{Overview of the method.} (\textbf{A}) Our training consists of two steps. First, we assume that the environment is fully observable and train a teacher state policy $\pi_s(a_t|s_t)$. Second, we train a student observation policy using a combination of teacher imitation and reinforcement learning. (\textbf{B}) We leverage fast GPU simulation powered by Isaac Gym and parallelize training across four A100 GPUs and thousands of randomized environments. Once a policy is trained in Isaac Gym, we validate it in the high-fidelity simulator provided by the robot manufacturer. Finally, we transfer it to the real robot. (\textbf{C}) Our neural network controller is a causal transformer model trained by autoregressive prediction of the next action from the history of observations and actions. We hypothesize that the observation-action history contains useful information about the world that a powerful transformer model can leverage to adjust its actions in-context.}
\label{fig:framework}
\end{figure}

In practice, estimating true underlying state of an environment is impossible for real-world applications. In the presence of a noisy observation space, the MDP framework needs to be modified to reflect the uncertainty in the observations. This can be done by introducing an observation space $O$ and an observation function $Z(o_t | s_t)$, which determines the probability of observing state $s_t$ as $o_t$. The MDP now becomes a Partially Observable Markov Decision Process (POMDP), where the agent must make decisions based on its noisy observations rather than the true state of the environment. 
The composition of the action, observation and state spaces is described in the following section. We illustrate our framework in Figure~\ref{fig:framework} and provide a comprehensive description of the method below.

\paragraph*{Model architecture.}
Our aim is to find a policy $\pi_o$ for real-world deployment in the POMDP problem. Our policy takes as input a history trajectory of observation-action pairs over a context window of length $l$, represented as ${o_t, a_{t-1}, o_{t-1}, a_{t-2}, ..., o_{t-l+1}, a_{t-l}}$, and outputs the next action $a_t$. To achieve this, we utilize transformers~\cite{Vaswani2017} for sequential trajectory modeling and action prediction.

Transformers are a type of neural network architecture that have been widely used in sequential modeling tasks, such as natural language processing~\cite{Devlin2018,Radford2018,Brown2020}, audio processing~\cite{Dong2018}, and increasingly in computer vision~\cite{Carion2020,Dosovitskiy2021} as well. The key feature of transformers is the use of a self-attention mechanism, which allows the model to weigh the importance of each input element in computing the output. The self-attention mechanism is implemented through a self-attention function, which takes as input a set of queries $Q$, keys $K$, and values $V$ and outputs a weighted sum, computed as follows:
\begin{equation}
    \mathrm{Attention}(Q,K,V)=\mathrm{softmax}(\frac{QK^T}{\sqrt{d_k}})V,
\end{equation}
where $d_k$ is the dimensionality of the key. The self-attention mechanism enables the transformer to capture long-range dependencies between input elements.

We represent each observation-action pair in the locomotion trajectory as a token. Transformers are able to extract the structural information of these tokens through a repeated process of assigning weights to each token (softmax on $Q$ and $K$) in time, and mapping the tokens ($V$) into features spaces, effectively highlighting relevant observations and actions and thus enabling the inference of important information such as gait and contact states. We employ Multi-Layer Perceptrons (MLPs) to embed each observation-action pair into a feature space. To capture the positional information of each token in the sequence, we add sinusoidal positional encodings to the features. We leverage the temporal dependencies among the observations and actions by restricting the self-attention mechanism to only attend to preceding tokens, resulting in a causal transformer~\cite{Radford2018}.

Transformers have proven to be effective in the realm of in-context learning, where a model's behavior can be dynamically adjusted based on the information present in its context window. Unlike gradient-based methods that require fine-tuning on task-specific data samples, transformers can learn in-context, providing them with the flexibility to handle diverse inputs.

The transformer model used in this study has four blocks, each of which has an embedding dimension of 192 and employs a multi-head attention mechanism with 4 heads. The MLP ratio of the transformer is set to 2.0. The hidden size of the MLP for projecting input observations is [512, 512]. The action prediction component of the model uses an MLP with hidden sizes of [256, 128]. Overall, the model contains 1.4M parameters. We use a context window of 16. The teacher state model is composed of an MLP with hidden sizes of [512, 512, 256, 128].

\paragraph*{Teacher state-policy supervision.}
In Reinforcement Learning (RL), an agent must continuously gather experience through trial-and-error and update its policy in order to optimize the decision-making process. However, this process can be challenging, particularly in complex and high-dimensional environments, where obtaining a useful reward signal may require a significant number of interactions and simulation steps. Through our investigation, we found that directly optimizing a policy using RL in observation space is slow and resource-intensive, due to limited sample efficiency, which impairs our iteration cycles.

To overcome these limitations, we adopt a two-step approach. First, we assume that the environment is fully observable and train a teacher state policy $\pi_s(a_t|s_t)$ using simulation. This training is fast and resource-efficient, and we tune the reward functions, such as gait-parameters, until an optimal state policy is obtained in simulation. Next, we distill the learned state policy to an observation policy through Kullback-Leibler (KL) divergence.

\paragraph*{Joint optimization with reinforcement learning.}
The discrepancy between the state space and the observation space can result in suboptimal decision-making if relying solely on state-policy supervision, as policies based on these separate spaces may have different reward manifolds with respect to the state and observation representations. To overcome this issue, we utilize a joint optimization approach combining RL loss with state-policy supervision. The objective function is defined as:
\begin{equation}
L(\pi_o) = L_{RL}(\pi_o) + \lambda D_{KL}(\pi_o \parallel \pi_s),\label{eq:joint}
\end{equation}
where $\lambda$ is a weighting factor representing the state-policy supervision, $L_{RL}(\pi_o)$ is the RL loss, and $D_{KL}(\pi_o \parallel \pi_s)$ is the KL divergence between the observation policy $\pi_o$ and the state policy $\pi_s$. The weighting factor $\lambda$ is gradually annealed to zero over the course of the training process, typically reaching zero at the mid-point of the training horizon, which enables the observation policy to benefit from the teacher early on and learn to surpass it eventually. It is important to note that our approach does not require any pre-computed trajectories or offline datasets, as both the state-policy supervision and RL-supervision are optimized through on-policy learning.

We use the proximal policy optimization (PPO)
algorithm~\cite{Schulman2017} for training RL policies.
The hyperparameters used in our experiments are shown in the supplement.
We use the actor-critic method and do not share weights.
The supplement lists the composition of the state and observation spaces. The action space consists of the PD setpoints for 16 actuated joints and the predicted PD gains for 8 actuated leg joints.
We do not train the policy to control the four toe motors, and instead we set the motors as their default positions using fixed PD gains. This is a widely adopted approach in model-based control \cite{Da2016, Gong2019}.

Our reward function is inspired by biomechanics study of human walking and tuned through trial and error. We do not have pre-computed gait library in our reward design. The detailed composition of our reward function can be found in the supplement.

\subsection*{Simulation}

\paragraph*{Closed kinematic chain.} 
In our simulation environment, we use the Isaac Gym simulator~\cite{Makoviychuk2021, Rudin2021} to model the rigid-body and contact dynamics of the Digit humanoid robot. Given the closed kinematic chains and underactuated nature of the knee-shin-tarsus and tarsus-toe joints of the robot, Isaac Gym is unable to effectively model these dynamics. To address this limitation, we introduce a ``virtual spring'' model with high stiffness to represent the rods. We apply forces calculated from the spring's deviation from its nominal length to the rigid bodies. Additionally, we employ an alternating simulation sub-step method to quickly correct the length of the virtual springs to their nominal values. We found that these efforts collectively make sim-to-real transfer feasible.

\paragraph*{Domain randomization.} 
We randomize various elements in the simulation, including dynamics properties of the robot, control parameters, and environment physics, as well as adding noise and delay to the observations. The supplement summarizes the domain randomization items and the corresponding ranges and distributions.
For the robot's walking environment, we randomize the terrain types, which include smooth planes, rough planes, and smooth slopes. The robot executes a variety of walking commands such as walking forward, sideward, turning, or a combination thereof, which are randomly resampled at a fixed interval. 
We set the commands below a small cut-off threshold to zero. The supplement lists the ranges of the commands used in our training.

\subsection*{Sim-to-real transfer}

The sim-to-real transfer pipeline is shown in Figure~\ref{fig:framework}.
We begin by evaluating our approach in the high fidelity Agility simulator developed by Agility robotics. This enables us to evaluate unsafe controllers and control for factors of variations. Unlike the Isaac Gym simulator that was used for training, Agility simulator accurately simulates the dynamics and physical properties of the Digit robot, including the closed kinematic chain structure that is not supported by Isaac Gym. In addition, Agility simulator simulates sensor noises characterized for the real Digit robot. Note that the policy evaluation in Agility simulator does not make any change to the neural network parameters. This step only serves to filter out unsafe policies.

For the deployment on hardware, we run the neural network policy at 50 Hz and the joint PD controller at 1 kHz. We can get access to joint encoders and IMU information through the API provided by Agility Robotics. We found that a combination of dynamics, terrain, and delay randomization leads to a high-quality sim-to-real transfer.

Finally, since the Isaac Gym simulator does not support accurate simulation of under-actuated systems, it poses additional challenges for sim-to-real transfer. In this study, we employed approximation methods to represent the closed kinematic chain structure. We believe that our framework will benefit from improving the simulator in the future.

\subsection*{Ablation studies}

\begin{figure}
\centering
\includegraphics[width=1.0\linewidth]{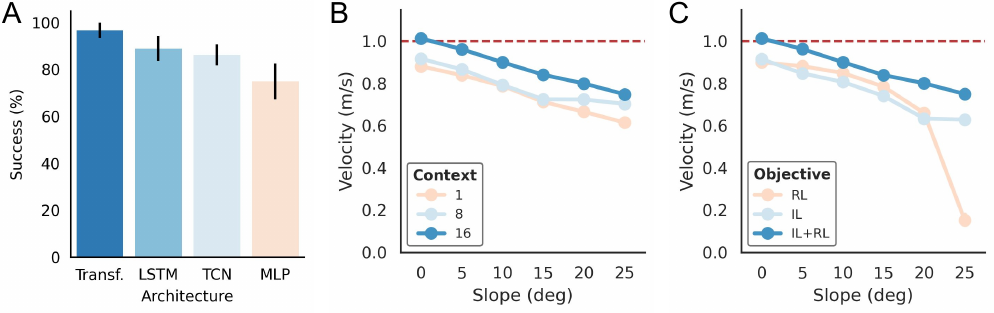}
\caption{\textbf{Ablation studies.} We perform ablation studies to understand the impact of key design choices. For fair comparisons, we keep everything fixed except for the varied component and follow the same hyper-parameter tuning procedure. (\textbf{A}) We find that the transformer models outperform the alternate neural network choices. (\textbf{B}) Our transformer-based controller benefits from larger context lengths. (\textbf{C}) Training with the joint objective consisting of both the imitation and reinforcement learning terms outperforms training with either of the two alone.}
\label{fig:ablations}
\end{figure}

In this section we preform ablation studies to analyze the key design choices in method. We compare different neural network architectures, context lengths, and training objective variants. Moreover, we analyze the attention maps of our transformer controller.

\paragraph*{Neural network comparisons.} We consider four different neural network architectures: 1) a Multi-Layer Perceptron (MLP), 2) a Temporal Convolutional Network (TCN) \cite{Bai2018}, 3) a Long Short-Term Memory (LSTM) \cite{Hochreiter1997} and 4) a Transformer model~\cite{Vaswani2017}. The MLP is widely used for quadrupedal locomotion \cite{Tan2018, Rudin2021}. The TCN achieves state-of-the-art quadrupedal locomotion performance over challenging terrain~\cite{Lee2020}. The LSTM shows the state-of-the-art performance for bipedal locomotion \cite{Siekmann2021sim, Siekmann2021blind}. Transformer models have not been used for humanoid locomotion before but have been incredibly impactful in natural language processing~\cite{Brown2020}. For fair comparisons, we use the same training framework for all neural network architectures and vary only the architecture of the student policy (Figure~\ref{fig:framework}). We optimize the hyper parameters for each of the models separately, control for different network sizes, and pick the settings that performs the best for each model choice.

In Figure~\ref{fig:ablations}A, we report the mean success rate and the 95\% confidence interval (CI) computed across 30 trials from 3 different scenarios from Figure~\ref{fig:indoors}D. We find that the transformer model outperforms other neural network choices by a considerable margin. Given the scaling properties of transformer models in NLP~\cite{Kaplan2020}, this is a promising signal for using transformer models for scaling learning-based approaches for real-world humanoid locomotion in the future.

\paragraph*{Transformer context length.} A key property of our transformer-based controller is to adapt its behavior implicitly based on the context of observations and actions. In Figure~\ref{fig:ablations}B, we study the performance of our approach for different context lengths. We command the robot to work forward at 1 m/s over two different slopes. We randomize the initial positions and heading and report the mean linear velocity and 95\% CI across 20 trials. We find that our model benefits from a larger context length in both settings.

\paragraph*{Training objective.} Our training objective from Equation~\ref{eq:joint} consists of two terms, an imitation learning term based on teacher policy supervision and a reinforcement learning term based on rewards. We study the impact of both terms. Using only the imitation term is common in quadrupedal locomotion~\cite{Lee2020} while using only reinforcement learning term corresponds to learning without a teacher~\cite{Openai2018, Siekmann2021sim}. In Figure~\ref{fig:ablations}C, we report the results on the same slope setting as in the previous context length ablation. We find that the joint imitation and reinforcement learning objective outperforms using either of the two terms alone.

\section*{Acknowledgments}

This work was supported in part by DARPA Machine Common Sense program, ONR MURI program (N00014-21-1-2801), NVIDIA, InnoHK of the Government of the Hong Kong Special Administrative Region via the Hong Kong Centre for Logistics Robotics, The AI Institute, and BAIR’s industrial alliance programs. We thank Sarthak Kamat, Baifeng Shi, and Saner Cakir for help with experiments; Aravind Srinivas, Agrim Gupta, Ashish Kumar, William Peebles, Tim Brooks, Matthew Tancik, Shuxiao Chen, Zhongyu Li, and Benjamin McInroe for helpful discussions; Gavriel State, Philipp Reist, Viktor Makoviychuk, Ankur Handa, and the Isaac Gym team for simulation discussions; Jack Thomas, Jake Thompson, Levi Allery, Jonathan Hurst, and the Agility Robotics team for hardware discussions.

\bibliography{scibib}
\bibliographystyle{IEEEtran}

\end{document}